# Salient Structure Detection by Context-Guided Visual Search

Kai-Fu Yang, Hui Li, Chao-Yi Li, and Yong-Jie Li*

University of Electronic Science and Technology of China, Chengdu, China.
*yang_kf@163.com, liyj@uestc.edu.cn*

**Abstract**—We define the task of salient structure (SS) detection to unify the saliency-related tasks like fixation prediction, salient object detection, and other detection of structures of interest. In this study, we propose a unified framework for SS detection by modeling the two-pathway-based guided search strategy of biological vision. Firstly, context-based spatial prior (CBSP) is extracted based on the layout of edges in the given scene along a fast visual pathway, called non-selective pathway. This is a rough and non-selective estimation of the locations where the potential SSs present. Secondly, another flow of local feature extraction is executed in parallel along the selective pathway. Finally, Bayesian inference is used to integrate local cues guided by CBSP, and to predict the exact locations of SSs in the input scene. The proposed model is invariant to size and features of objects. Experimental results on four datasets (two fixation prediction datasets and two salient object datasets) demonstrate that our system achieves competitive performance for SS detection (i.e., both the tasks of fixation prediction and salient object detection) comparing to the state-of-the-art methods.

## 1. Introduction

Visual search is necessary for rapid scene analysis because information processing in the visual system is limited to one or a few regions at one time [3]. To select potential regions or objects of interest rapidly with a task-independent manner, the so-called "visual saliency", is important for reducing the complexity of scenes. From the perspective of engineering, modeling visual saliency usually facilitates subsequent higher visual processing, such as image re-targeting [10], image compression [12], object recognition [16], *etc*.

Visual attention model is deeply studied in recent decades. Most of existing models are built on the biologically-inspired architecture based on the famous *Feature Integration Theory (FIT)* [19, 20]. For instance, Itti *et al.* proposed a famous saliency model which computes the saliency map with local contrast in multiple feature dimensions, such as color, orientation, etc. [15] [23]. However, FIT-based methods perhaps risk being immersed in local saliency (e.g., object boundaries), because they employ local contrast of features in limited regions and ignore the global information. Visual attention models usually provide location information of the potential objects, but miss some object-related information (e.g., object surfaces) that is necessary for further object detection and recognition.

Distinguished from FIT, *Guided Search Theory (GST)* [3] [24] provides a mechanism to search the regions of interest (ROI) or objects with the guidance from scene layout or top-down sources. The recent version of GST claims that the visual system searches objects of interest along two parallel pathways, i.e., the non-selective pathway and the selective pathway [3]. This new visual search strategy allows observers to extract spatial layout (or gist) information rapidly from entire scene via non-selective pathway. Then, this context information of scene acts as top-down modulation to guide the salient object search along the selective pathway. This two-pathway-based search strategy provides a parallel processing of global and local information for rapid visual search. Referring to the GST, we assume that the non-selective pathway provides "where" information and prior of multiple objects for visual search, a counterpart to visual selective saliency, and we use certain simple and fast fixation prediction method to provide an initial estimate of where the objects present. At the same time, the bottom-up visual selective pathway extracts fine image features in multiple cue channels, which could be regarded as a counterpart to the "what" pathway in visual system for object recognition. When these bottom-up features meet "where" information of objects, the visual system achieves object detection and further for object recognition.

In this paper, we propose a novel *Context-Guided Visual Search (CGVS)* model to implement the two-pathway-based visual search strategy [3], but with a task-independent manner. The global context-based spatial prior is represented by the distribution of dominant edges of scene via non-selective pathway. The layout representations from non-selective pathway are used as the initial guidance to evaluate the location and size of regions of interest and the relative importance of local cues. On the other hand, the local cues (including color, luminance, and texture, etc.) are extracted in parallel along selective pathway. Finally, we use Bayesian inference to integrate the global layout information and local cues, and to predict the saliency of each pixel. The salient structures are further enhanced via an iterative processing to re-initialize the prior guidance as the final prediction.

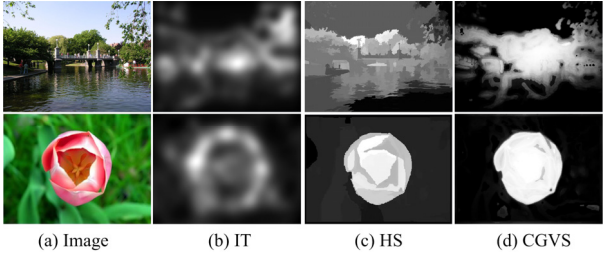

Fig. 1. Comparing with the tasks of fixation prediction (b) and salient object detection (c), our salient structure detection (d) aims to extracting interesting structures for both complex and simple scenes.

In the experiments, we apply our model to *Salient Structure Detection* task. At the beginning, we define the salient structure detection as detecting the accurate regions containing structures of interest in a natural scene. This means that such task aims at identifying the regions of interest (ROIs) in complex scenes, while detecting dominant objects in simple scenes. Fig. 1 shows two examples of salient structure detection in natural scenes. Fixation prediction methods (e.g., IT [15]) usually focus on high-contrast boundaries, and ignore object surfaces and shapes (Fig. 1b). In contrast, object detection models (e.g., HS [6]) may be inefficient for ROI detection in complex scenes without dominant objects (Fig. 1c). The proposed method can efficiently extract accurate salient structures for both complex and simple scenes (Fig. 1d). Extensively experimental evaluations on several benchmark datasets demonstrate that our model can handle both fixation prediction and salient object detection well without specific tuning.

The proposed system attempts to bridge the gap between two highly related tasks, i.e., the human fixation prediction [25] and salient object detection [26], with a simple and general framework. Almost all fixation prediction methods only detect sparse local regions, and cannot well handle the task of salient object detection. In contrast, the proposed method can efficiently transform the saliency map of fixation prediction methods to clear salient objects.

To summarize the above, contributions of the proposed work are as follows. (1) We define a new task called salient structure detection to unify various saliency-related tasks like fixation prediction, salient object detection, and detection of other structures of interest. (2) We propose a unified framework for object search based on the *Guided Search Theory*. (3) The proposed system provides an efficient way for transforming saliency map to salient structures and multi-object search. (4) The comprehensive experiments on several benchmark dataset reveal the effectiveness of the proposed system for salient structure detection and object search using saliency map.

## 2. Related Work

In recent years, a large number of selective attention models are proposed to predict the human fixations in natural scenes[25]. Since the work of Itti *et al.* [15], lots of methods compute the saliency map using the local contrast in multiple scales and feature dimensions (e.g., color, orientation, etc.) with a bottom-up framework [25, 27]. In addition, other methods of saliency computation include Graph-based (GB) [11], Information Maximization (AIM) [17], Image Signature (SIG) [18], Adaptive Whitening Saliency (AWS) [22], and Local and Global Patch Rarities [28]. Frequency domain based models include Spectral Residual (SR) [29], Phase spectrum of Quaternion Fourier Transform (PQFT) [30, 31], and Hypercomplex Fourier Transform (HFT) [8].

On the other hand, machine learning techniques are usually introduced to improve the performance of fixation prediction. In these models, both bottom-up and top-down visual features are learnt to predict salient locations [1, 32]. In general, interesting objects (like humans, faces, cars, text, and animals, etc.) convey more information in a scene, and they usually attract more human gaze [33-36]. Task-related top-down information is also commonly used to facilitate specific object search [37-40]. Some models also learn optimal weights for channel combination in the bottom-up architecture [41], and nonparametric saliency models learn directly from human eye movement data [42].

Other models attempt to improve visual search in complex scenes by employing statistics features based on probabilistic formulation. Zhang *et al.* [21] proposed a Bayesian framework for saliency computation using natural statistic. Torralba *et al.* [43, 44] used the global features to guide object search by summarizing the probability regions of presence of target objects in the scene. Itti *et al.* [45] proposed the Bayesian definition of surprise by measuring the difference between posterior and prior beliefs of the observer. More discussion about selective attention models can be found in a recent review paper [25]. These models are usually validated against eye movement data recorded from human observers [8].

Saliency map from an attention model represents possible fixations but misses the object-related information (e.g. contours and surfaces of objects). In order to accurately extract the dominant objects from natural scenes, Liu *et al.* [46] formulated the salient object detection as a binary labeling problem. Achanta *et al.* [2] further claimed that salient object detection requires to label pixel-accurate object silhouette. Most of the existing methods have attempted to detect the most salient object based on local or global region contrast [6, 13, 14, 47]. For example, Cheng *et al.* [13] proposed a region-based method for salient object detection by measuring the global contrast between the target region and other regions. More methods include center-surround contrasts with Kullback-Leiblar [48], background prior [49, 50], *etc.* An overview of salient object detection methods can be found in a recent review [26]. Salient object detection also highly relates to another task called *object proposal*, which attempts to generate a set of all objects in the scene, regardless of the specific saliency of these objects [51-53].

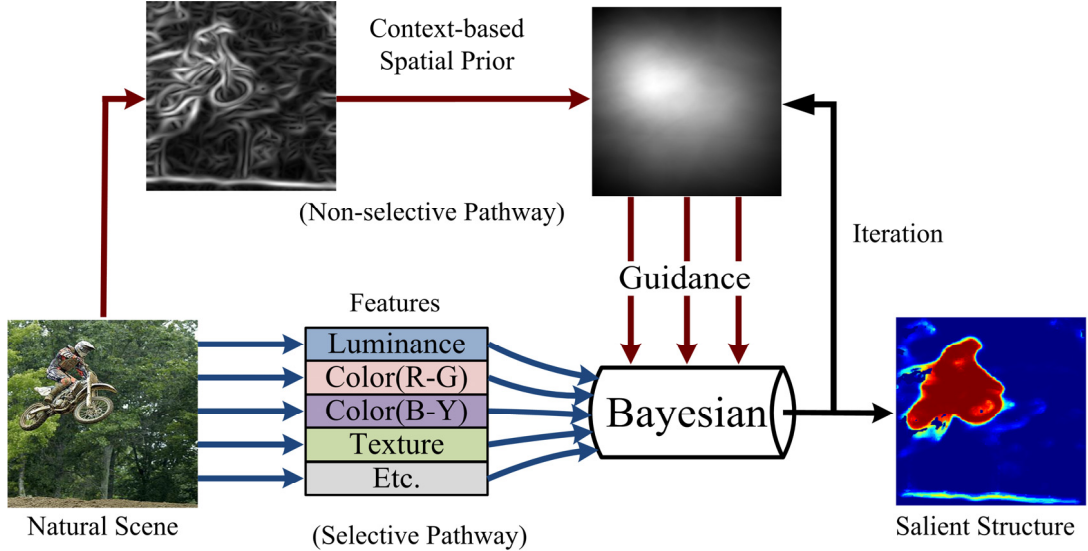

Fig. 2. The flowchart of the proposed system including the selective and non-selective pathways.

A related model is proposed by Xie, *et al*. [5], which also employs Bayesian framework for saliency computation with the prior informantion evaluated using the convex hull of Harris points and Laplacian sparse subspace clustering algorithm. Unlike this method, our method obtains the spatial prior informantion in a much simpler way. Meawhile, the context-based information in our model is used to identify the sizes of potential objects and the importance of local cues. In addition, the proposed visual search model provides a unified and bilogically-plausible framework for both fixation prediction and salient object detection, called as "Salient Structure Detection" in this paper (see Fig. 1).

More recently, several authors attemped to bridge the gap between the two tasks mentioned above. Goferman *et al*. [10] proposed a context-aware saliency algorithm to detect both prominent objects and the parts of the background that convey the context. However, their method still overemphasizes the boundary of object, i.e., high-contrast boundaries of objects often stand out instead of the surface of objects. Li *et al*. [54] trained a random regreession forest to extract salient regions by employing a state-of-the-art image segmentation method. In comparison to [10], our method obtains more reasonable salient structures with high-accuracy object silhouette and object surfaces. In addition, compared with [54], the proposed model can handle well for salient structure detection in both simple and complex scenes without foregone computation such as image segmentation.

A method based on *Boolean Map* was proposed for fixation prediction [7]. That method can also be used for salient object detection by adding a specific tuning and post-processing. In contrast, our method can achieve salient structure detection with the totally same system. In addition, the proposed framework is a bilogically-plausible architecture of visual search, which has potential superiority in explaining the information processing in visual search.

## 3. Context-Guided Visual Search Model

The flowchart of the proposed method is summarized in Fig. 2. Dominant edges contain main layout information of a natural scene by dividing the whole scene into several perceptional regions. In the proposed system, the possible locations of potential salient structures are evaluated with the distribution of dominant edges in the non-selective pathway. Meanwhile, the features like color, luminance, and texture are extracted from the given scene in the selective pathway. The context-based spatial prior information is fed into the Bayesian framework for feature integration and salient structure prediction. Finally, the output of Bayesian framework is used as new spatial prior to re-evaluate the salient structure detection with an iterative process.

### 3.1 Context-based Spatial Prior

In non-selective pathway, we compute the rough spatial weights of saliency based on the distribution of the dominant edges. In fact, edge information has been widely used for saliency computation. For example, Itti *et al*. [50] considered the edge as orientation features and other methods use edge density as the evaluation of saliency [55]. However, these methods cannot provide region information (e.g., object surfaces), while some isolated and high-contrast edges (e.g., the boundary of two large surfaces) may be incorrectly evaluated as high saliency.

In this paper, we try to (roughly) reconstruct potential saliency regions based on the information of dominant edges. In detail, we first extract the edge responses and the corresponding orientations using the edge detector proposed in [56], which is a biologically-inspired method and can efficiently detect both color and brightness defined boundaries from natural scenes. Fig. 3 shows an example of reconstructing potential saliency regions utilizing dominant edges. For each edge pixel, we compute the average edge response in the "left" and "right" half disk around it. The disk is defined by the orientation of each edge pixel, with a radius of $d_r$. In this work, we experimentally set $d_r = min(W, H)/3$, where $W$ and $H$ indicate the width and height of the given image, respectively. Then, the all pixels within the half disk with higher edge response are voted 1, and the pixels in another half disk are voted 0. For each pixel, its saliency weight is represented by the number of votes when all edges pixels finish their votes. In the experiment, we just scan the dominant edges (i.e., the ridges, red pixels in Fig. 3(left)) to speed up the computation. Fig. 3 (right) shows the possible regions where salient structures present.

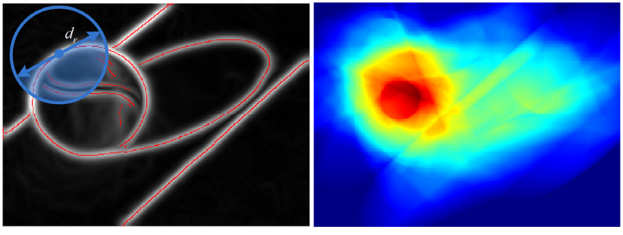

Fig. 3. Example of reconstructing potential saliency regions by utilizing dominant edges. **Left**: the dominant edges are shown in red lines. For each edge pixel, the averaged edge responses in the "left" and "right" half disks around it are summarized to decide which half is located in the salient region. **Right**: the possibility of locations where the salient structure presents.

We denote the rough spatial weights of saliency as $S_e$. In addition, we also consider the center-bias weights [1, 57] (denoted by $S_c$) modeled by a Gaussian function with the standard deviation as $\sigma_c = min(W, H)/3$. The values of $S_e$ and $S_c$ are linearly normalized to the range of [0, 1]. Then, the final Context-based Spatial Prior (CBSP) is given by

$$S_w = S_e + S_c \qquad (1)$$

Here we take the form of summation of $S_e$ and $S_c$, rather than the form of multiplication that is used in most previous work (e.g., [10]), because we use both $S_e$ and $S_c$ as spatial priors of equal importance. In addition, $S_w$ is also linearly normalized to the range of [0,1] for being used later as prior probability.

### 3.2 Feature Extraction

In the selective pathway, low-level features including color, luminance and texture are parallel extracted. With $r$, $g$, and $b$ denoting the red, green, and blue components of the input color image, luminance channel is obtained as $f_{lum} = (r+g+b)/3$, and two color-opponent channels are computed with $f_{rg} = r-g$ and $f_{by} = b-(r+g)/2$, respectively. The luminance and two color-opponent channels are smoothed with a Gaussian filter with the scale same as that used in edge detector [56] to remove noises.

In addition, a texture channel ($f_{ed}$) is represented by edge density which is computed by smoothing the edge response (same as that used in the non-selective pathway, Section 3.1) with an average filter of $11 \times 11$ pixels.

### 3.3 Bayesian Inference with Context Guidance

Following the visual search strategy, context-based information is used to guide the integration of local features to compose a salient structure. In this paper, we employ the tool of Bayesian inference to adaptively integrate the global layout and local feature information, simulating the interaction of top-down and bottom-up information processing flows in selective visual attention.

With Bayesian inference, the possibility of a pixel belonging to a salient structure (posterior probability) can be computed as

$$p(s|x) = \frac{p(s)p(x|s)}{p(s)p(x|s) + p(b)p(x|b)} \qquad (2)$$

where $p(s)$ and $p(b) = 1 - p(s)$ are respectively the prior probabilities of a pixel belonging to a salient structure and the background. $p(x|s)$ and $p(x|b)$ are likelihood functions based on the observed salient structure and the background, respectively. In this work, we set CBSP extracted from the non-selective pathway (Equation (1)) as the initial prior probability, i.e., $p(s) = S_w$. Meanwhile, the likelihood functions will be evaluated adapting to each scene context, including the possible sizes of salient structures and the relative importance of each feature.

**Predict the size of potential structure.** To obtain likelihood functions of observed objects and background, we first extract the possible regions containing structures from background. Simply, we binarize the map of prior probability ($p(s)$) with an adaptive threshold to capture rough potential structures and their sizes.

We use $S_{T_k}$ and $B_{T_k}$ to denote the sets of structure and background pixels obtained by binarizing $p(s)$ with certain threshold $T_k$. The optimal threshold $T_{max}$ is found by searching a possible $T_k$ which maximizes the difference of all features between structure and background pixel sets according to

$$T_{max} = arg \max_{T_k} \sqrt{\sum_i \left(\omega_i^0 \cdot \left(\tilde{S}_{T_k}^i - \tilde{B}_{T_k}^i\right)\right)^2} \qquad (3)$$

where $\tilde{S}_{T_k}^i$ and $\tilde{B}_{T_k}^i$ denote respectively the average values of structure and background pixels in feature channel $i$, $i \in \{f_{lum}, f_{rg}, f_{by}, f_{ed}\}$. The initial feature weight is $\omega_i^0 = 0.25$, which indicates the equal importance of each cue at the initial status. $T_k \in \{10\%, 12\%, 14\%, ..., 50\%\}$ indicates the percentage of pixels of the potential salient structures. This suggests a potential assumption that salient structures are usually smaller than half of the image.

Meanwhile, we ignore the regions with fine scales (<10%) to avoid fragments. This assumption is supported by a simple experiment on two popular salient object datasets: the mean sizes (percentage) of salient objects are 20.01% on ASD [2] and 28.51% on ECSSD datasets [6].

**Evaluate the importance of each feature.** After finding the potential salient and background pixel sets for each feature map based on the optimal threshold $T_{max}$, we re-evaluate the importance of each feature as

$$\omega_i = \frac{1}{\mu} \cdot \left| \tilde{S}^i_{T_{max}} - \tilde{B}^i_{T_{max}} \right| \quad (4)$$

where $\mu = \sum_i \omega_i$, $i \in \{f_{lum}, f_{rg}, f_{by}, f_{ed}\}$. Equation (4) indicates that a feature will have higher importance when the difference of averaged pixel values between structure and background is larger in this channel.

**Calculate the likelihood functions.** We then compute the likelihood functions with the potential object and background pixel sets and the weight of each feature. We assume that the four feature channels are independent, and the likelihood functions can be obtained as

$$p(x|s) = \prod_{i \in \{f_{lum}, f_{rg}, f_{by}, f_{ed}\}} p^{\omega_i}(x_i | S_{T_{max}}) \quad (5)$$

$$p(x|b) = \prod_{i \in \{f_{lum}, f_{rg}, f_{by}, f_{ed}\}} p^{\omega_i}(x_i | B_{T_{max}}) \quad (6)$$

where $p^{\omega_i}(x_i | S_{T_{max}})$ and $p^{\omega_i}(x_i | B_{T_{max}})$ are respectively the distribution functions of each feature ($i \in \{f_{lum}, f_{rg}, f_{by}, f_{ed}\}$) in the salient structure and background sets. $\omega_i$ indicates the contribution of distribution function of the $i$-th feature. Finally, the posterior probability ($p(s|x)$) is computed with Equation (2) as the saliency of each pixel.

**Enhance the salient structure by iterating.** We further enhance the salient structures iteratively by re-initializing the prior function with $p(s) \leftarrow p(s|x)$ and the feature weights as $\omega_i^0 \leftarrow \omega_i$. In the experiment, we re-initialize the prior function with the smoothed version of $p(s|x)$ (median filtering with size of $21 \times 21$) to remove some small fragments. Finally, we denote the $CGVS(t)$, $t = t_0, t_1, ..., t_n$ as our context-guided visual search model with various iterations ($t$), and the $CGVS(t_0)$ is the first output without iteration.

## 4. Transform Saliency Map to Salient Structure

In addition, our system is expected to bridge the gap between the tasks of fixation prediction and salient structure detection. In details, with the saliency map computed by the fixation prediction methods, important

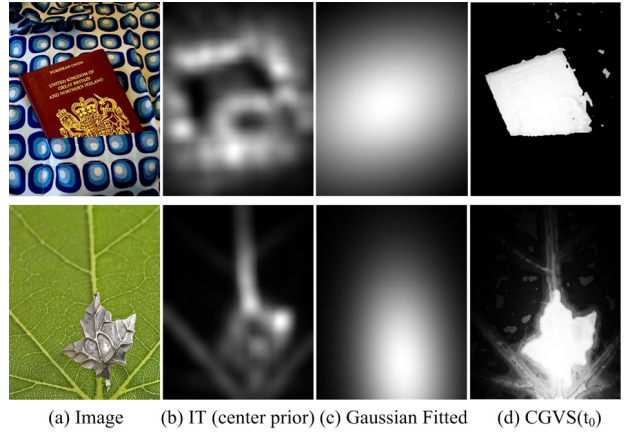

(a) Image    (b) IT (center prior)    (c) Gaussian Fitted    (d) CGVS($t_0$)

Fig. 4. The steps of transforming saliency map to salient structure with the proposed system.

information usually distribute around saliency regions. We employ the commonly used center prior to weaken the effect of saliency regions close to the image border. Then, we fit the global distribution of saliency map with a 2-Dimension Gaussian function. The fitted result is used as initial CBSP (Equivalent to $S_w$ in Equation (1)) in non-selective pathway. Then, our system can transform the saliency map to salient structures with Equation (2)~(6).

Figure 4 show the steps of salient structure transformation. We first compute the saliency map with certain fixation prediction method (e.g., IT [15] with center prior shown in Fig. 4b). The fitted Gaussian function is shown in Fig. 4c. Then the salient structure (Fig. 4d) is obtained by the proposed system with the Gaussian-fitted saliency map as the initial CBSP.

## 5. Experiments

In general, existing fixation prediction methods work to extract sparse salient locations, but fail to capture the fine structure of objects. In contrast, salient object detection methods usually detect dominant objects and are not suitable for the analysis of complex scenes without an obviously salient object. The proposed system aims to detect potential interesting structures including both the ROIs for complex scenes and the salient objects for simple scenes. Because there is no perfect benchmark dataset for salient structure detection, so the performance is evaluated on both fixation prediction datasets and salient object datasets in this experiment. The method obtaining higher scores on both of these two tasks can be considered as being more suitable for salient structure detection.

In this section, we will first show the basic properties of our system in scene analysis. Then the proposed method will be evaluated on both fixation prediction datasets [1, 8] and salient object detection datasets [2, 6]. We will also exploit the effect of parameters on detection performance and demonstrate that fixations prediction methods can be significantly improved for the task of salient object detection.

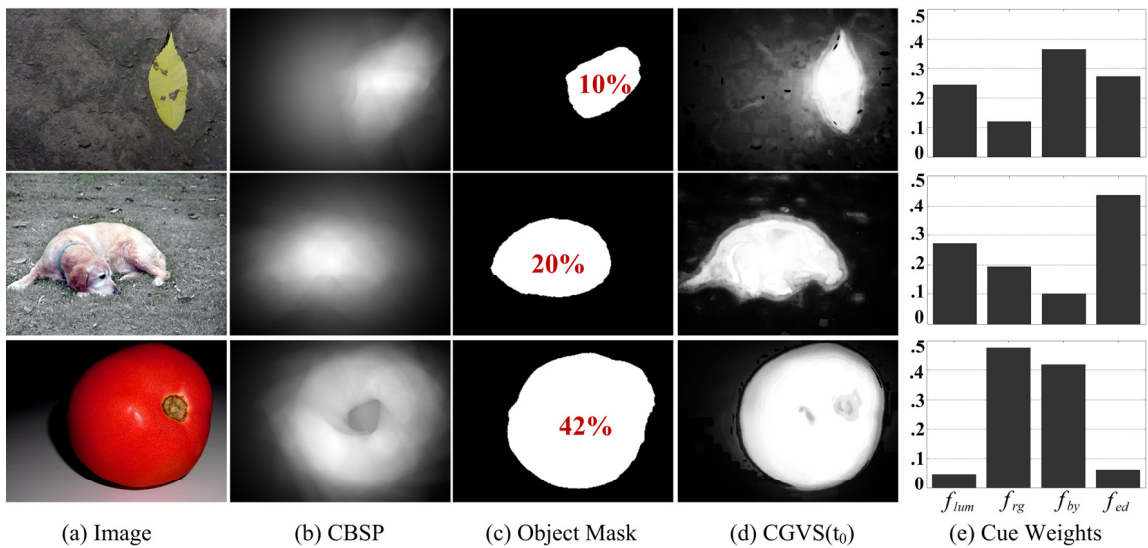

Fig. 5. Several examples illustrating that our system automatically selects the size of potential salient structure and estimates the importance of each local feature. The object mask in (c) is obtained by thresholding the CBSP in (b) with the currently optimal threshold (i.e., the percentage number shown in (c)).

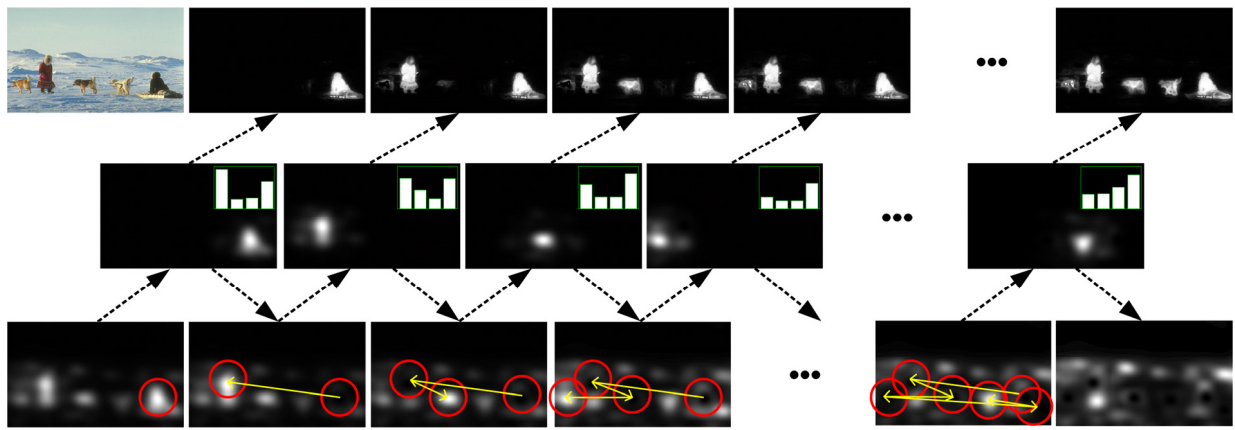

Fig. 6. The procedure of object searching in a multiple object scene.

**5.1. Basic Property of the Proposed Model**

To begin with, we evaluate the basic properties of the proposed model. The most important feature is that our model can automatically select the sizes of potential salient structures and estimate the relative importance of each local feature.

Figure 5 shows three examples including objects with different scales. With Equation (3), our system can automatically search the spatial sizes of potential objects in the given scene and roughly evaluate and identify the pixels of salient regions and background (Fig. 5c) by thresholding the CBSP (Fig. 5b) with certain threshold value. Then, the importance of each feature is computed with Equation (4), and the observed likelihood functions of salient regions and background are evaluated with Equation (5~6). Finally, the possibility of a pixel belonging to the salient structures (Fig. 5d) is obtained with Bayesian inference (Equation (2)). Fig. 5e lists the weights of all features. From Fig. 5, we can clearly see that our system obtains reasonable evaluation about the size of salient structure and the importance of features according to the input scene. This is important for searching task-free interesting structures from complex scenes.

An additional experiment was executed in order to model the process of object searching in multiple object scenes. Fig. 6 shows the object searching procedure beginning with an initial saliency map. Based on the previous work of Itti *et al*. [15], the mechanisms of "winner-take-all" and "inhibition-of-return" were employed to search salient locations. The bottom row of Fig. 6 shows the classical shift of focus of attention modeled by the method of Itti *et al*. [15]. Comparing to Itti's model, our system further extracts full structures (the top row) from each attended location (the middle row). The objects were found one by one with time course. In addition, our method provides the importance of each feature for each object, which is indicated by the histogram shown in Fig. 6 (the middle row). We believe these features are also useful for further object recognition.

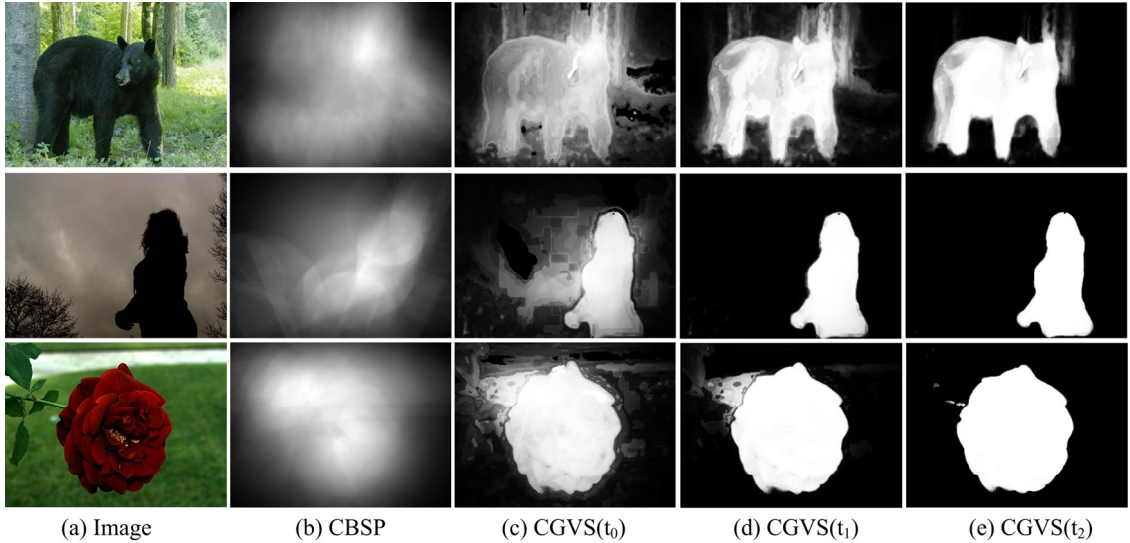

(a) Image      (b) CBSP      (c) CGVS($t_0$)      (d) CGVS($t_1$)      (e) CGVS($t_2$)

Fig. 7. Enhancing the salient structure in an iterative way. (a) Original image, (b) Context-based spatial prior (CBSP), (c)-(e) Results of our method at different iteration stages.

On the other hand, our model further re-evaluates salient structure with new prior and feature weights for improving the confidence of salient structures. Fig. 7 shows several examples which illustrate that our system can always correctly identify the salient object, although the initial CBSP (Fig. 7b) is inaccurate. For example, in the *bear* image (Fig. 7, the top row), the most salient location is on the head of bear and most parts of the bear's body are missed in the initial CBSP. However, after two steps of iteration, our model detects the full bear and suppresses background textures significantly.

**5.2. Fixation Prediction**
Fixation prediction methods are usually benchmarked on some available fixation datasets [1, 8]. Although our model detects the salient structures instead of the sparse fixations, it is believed that most human fixations should be present on the interest regions or salient structures in the complex scene. Therefore, we also evaluated the performance of our system with the benchmark of fixation prediction for complex scenes with ROC curve [1]. We evaluated the results of the proposed model on two large datasets collected by Judd, *et al*. [1] and Li, *et al*.[8] , which contain 1003 images and 235 images, respectively. In addition, we compared our method to several existing fixation prediction methods.

From Fig. 8, some existing methods usually provide stronger responses to regions with higher local contrasts, such as edges or boundaries of objects (e.g., CA[10], SUN[21]), while ignoring the surfaces of salient structures. Others models obtain highly blurred saliency map, which cannot provide fine shapes or structures of objects. In contrast, our CGVS method is efficient for various situations of scenes. For simple scenes with predominant objects, CGVS responds well to full objects (Fig. 8, the first to third rows). In addition, our method is also efficient when scenes contain multiple objects (two objects in Fig. 8, the fourth row) and large objects (Fig. 8, the fifth to sixth rows). In short, our CGVS saliency contributes to saliency computation in both simple and complex scenes.

We further quantitatively evaluated the performance of the proposed method for the task of fixation prediction. Fig. 9 (left) shows the ROC curves of CGVS with various iteration steps on two datasets. In general, CGVS($t_0$) achieves the best performance for fixation prediction. This is because that iteration processing makes our system focus on the most few salient objects in scenes, and against precision of fixation prediction. Fig. 9 (right) shows that our method outperforms (at least catches up) all considered bottom-up (low-level) methods. Note that the method proposed by Judd *et al*. [1] achieves better performance mainly because that a training process and several high-level features, such as face detection, person detection, *etc* are introduced into their model.

It is worth to note that the measure of ROC Curve is somewhat biased when evaluating the performance of fixation prediction. Just as that indicated by Goferman, *et al*.[10], incorporating a center prior to the final saliency estimation can remarkably improve quantitative evaluation, but makes the saliency map look less convincing visually. Fig. 10 shows several examples indicating that CA with center prior obtains high performance in ROC curve (Fig. 10, the last column), but misses lots of object information when visualizing the saliency maps (Fig. 10, third column). On the other hand, CA without center prior provides better results with qualitative evaluation (Fig. 10, the second column), but lower score on ROC curve.

Actually, most of existing methods promote higher saliency values in the center of the image plane, such as GBVS [11], Judd [1], etc. The proposed method also combines the commonly used center prior when computing CBSP. With similar observation with [10], CBSP achieves good performance on ROC curve with the very blurred saliency region (Fig. 10, the forth column) regardless of the structure information of objects. However, our final CGVS is usually capable of detecting full objects and surfaces

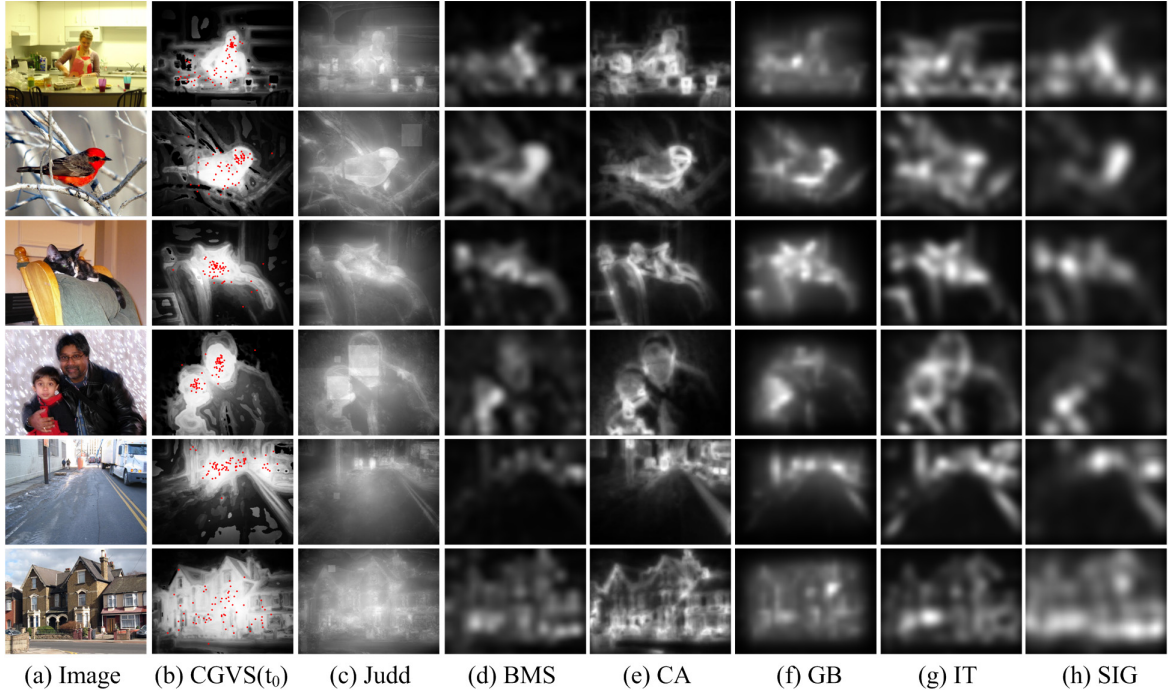

(a) Image  (b) CGVS($t_0$)  (c) Judd  (d) BMS  (e) CA  (f) GB  (g) IT  (h) SIG

Fig. 8. Comparing the fixation prediction results. (a) original images, (b) saliency structure produced using the proposed method (CGVS) without the iterative processing, saliency maps produced using multiple methods: (c) Judd et al. (Judd) [1], (d) Boolean map (BMS) [7], (e) Context-aware (CA) [10], (f) Graph-Based (GB)[11], (g) Itti et al. (IT) [15], and (h) Image Signature (SIG)[18]. Note that the red points overlapped on the CGVS maps indicate the human fixations (ground truth). It is clear that our CGVS generates uniformly highlighted salient structure that covers almost all human fixations.

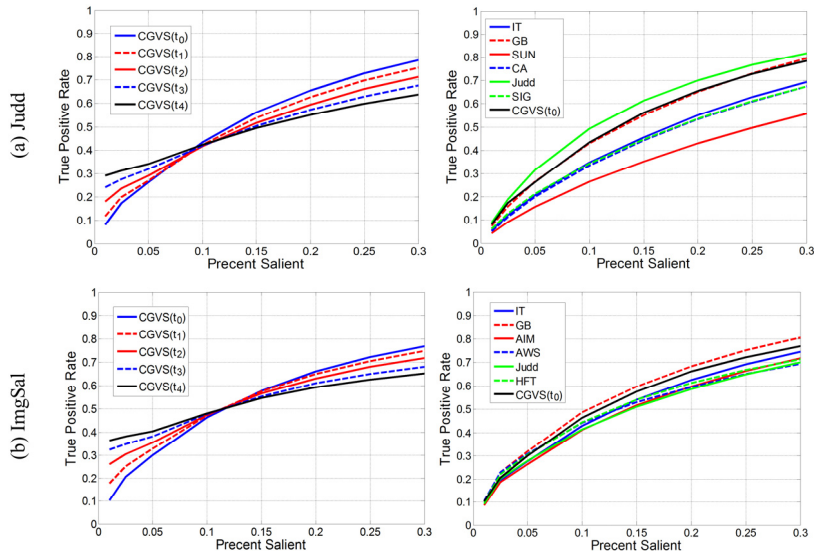

Fig. 9. Quantitative evaluation on two fixation prediction datasets (a) Judd [1] and (b) ImgSal [8]. **Left**: ROC curves for our method with various iteration steps. It is clear that CGVS($t_0$) achieves the best performance for fixation prediction. **Right**: ROC curves for our method comparing to other algorithms, indicating that our model outperforms the most methods and is comparable to GB[11]. The model of Judd *et al.* provides the best performance mainly because it uses learning and high-level object detection. Considered methods include IT[15], GB [11], AIM[17], SUN[21], AWS[22], SIG[18], CA[10], HFT[8], Judd[1].

(Fig. 10, the fifth column), and holding high performance on ROC at the same time (Fig. 10, the last column). Comparing to CA with or without center prior, CGVS achieves qualitatively better results when visualizing the saliency maps, although CA with center prior achieves higher performance on ROC curve. The reason is that CBSP is just a rough estimate of potential saliency regions, and the saliency of each pixel is reassigned with Bayesian Inference during the next step of iteration. Therefore, center prior implemented in CBSP will not cause the loss of saliency structure near the border of images.

In addition, there are several other metrics for saliency evaluation, and some of them (e.g., shuffled AUC [21]) are expected to tackle the influence of center prior [58].

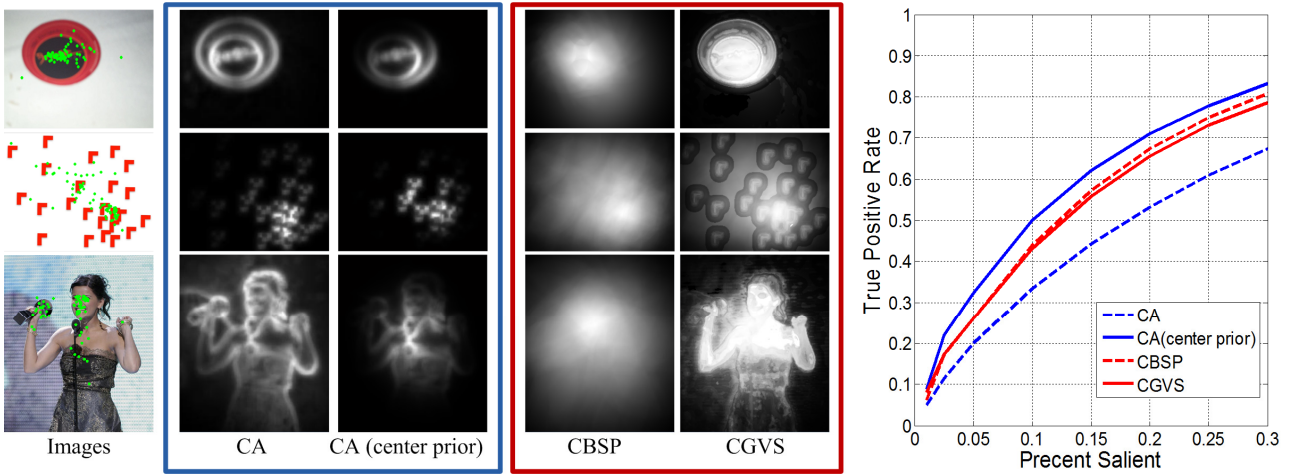

Fig. 10. Examples showing the gap between qualitative and quantitative evaluation. Quantitatively, CA provides large improvement when introducing center prior, but the visual assessment drops a lot. Similarly, although quantitative evaluation of our method is a little worse than that of CA with center prior, our CGVS obtains the excellent assessment when visualizing the saliency map, which highlights almost all the pixels of the dominant objects.

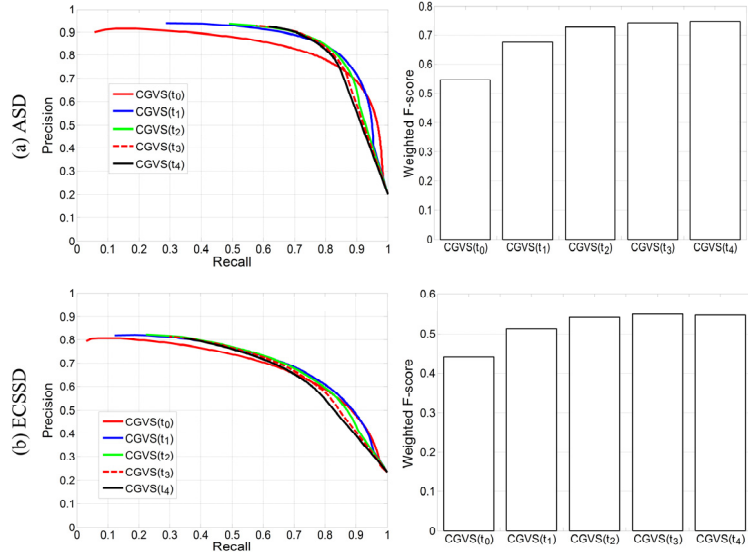

Fig. 11. Quantitative evaluation on two salient object datasets (a) ASD [2] and (b) ECSSD [6]. **Left**: P-R curves for our method with various iteration steps. **Right**: weighted F-score for our method with various iteration steps. CGVS($t_2$) achieves the stable performance for salient object detection.

However, all these metrics (including ROC) for evaluating fixation prediction are computed against the human fixations which are extremely sparse. The salient structures like some object surfaces extracted by our method may be inaccurately considered as false alarm when computing metrics, though we believe that extracting object information is a main merit of our method.

### 5.3. Salient Object Detection

For simple scenes with dominant objects, a lot of algorithms have been successfully developed for salient object detection. Unlike fixation prediction, salient object detection methods are commonly benchmarked by binary pixel-accurate object masks [2].

In this experiment, we first use the standard F-measure (*P-R curve* and *F-score*) for performance evaluation on two popular datasets: ASD [2] and ECSSD [6]. However, Along the systemic analysis of Margolin *et al*. [59], F-measure does not always provide a reliable evaluation for salient object detection. Therefore, we also employ the amended F-measure (called *Weighted F-score*) proposed in [59] as a more preferable measure. In addition, another measure of *Mean Absolute Error (MAE)* [60] [14, 49] is also introduced for complementary performance evaluation.

Fig. 11 lists the performance of our methods with both the standard P-R curves and the weighted F-scores on ASD and ECSSD datasets. Different from fixation prediction in Section 5.2, the iterating process can further enhance the regions of objects and improve the performance since the benchmark used in this experiment has few salient objects. Generally, our CGVS system can obtain a stable

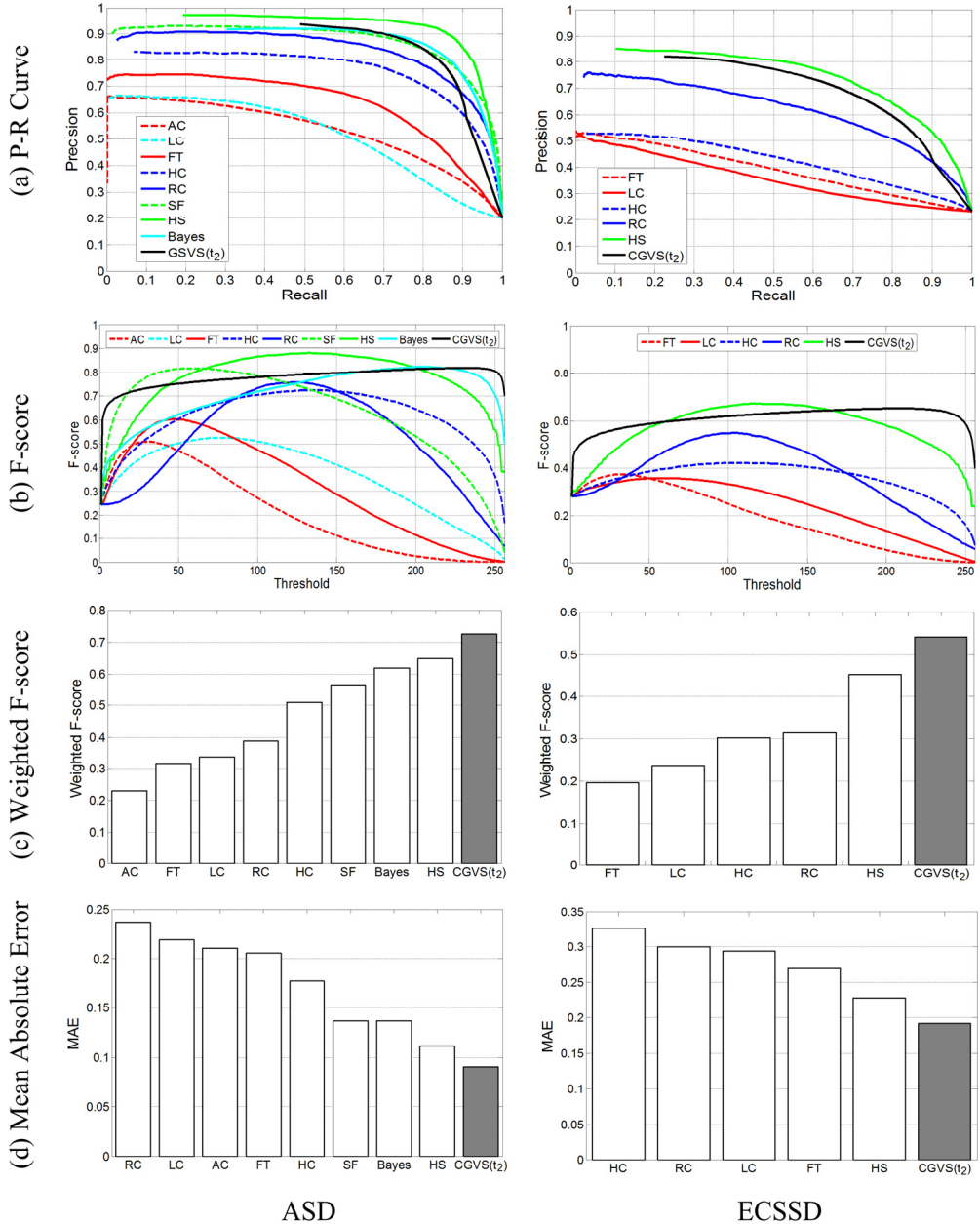

Fig. 12. Statistical comparison of different saliency detection methods with four metrics on two datasets. (a) P-R Curve, (b) F-score, (c) Weighted F-score, and (d) Mean Absolute Error (MAE). The methods considered include AC[4], LC[9], FT[2], HC[13], RC[13], SF[14], HS[6], Bayes[5].

performance with only two times of iteration (i.e., CGVS($t_2$)).

Fig. 12 shows *P-R Curve*, *F-score*, *Weighted F-score*, and *MAE* for various state-of-the-art salient object methods on the two datasets. From the standard P-R curve and F-score, our CGVS($t_2$) outperforms most of the considered algorithms except HS [6]. However, our method achieves high F-score across a large range of thresholds (Fig. 12, the second row) on both datasets. This result indicates that the proposed model is capable of obtaining high confidence of salient objects. In addition, our method significantly outperforms all considered algorithms on the other measures of weighted F-score and MAE. Totally, our system achieves competitive performance comparing to the state-of-the-art methods of salient object detection in simple scenes.

Several example results of salient object detection are shown in Fig. 13.

### 5.4 Robustness to Parameters
In the proposed system, the observed object mask and cue weights are important for the evaluation of likelihood function in Section 3.3. Fortunately, our system is capable of automatically predicting the size of potential structure and the relative importance of each feature according to CBSP (Equation (1)). Therefore, the other two parameters in the computation of CBSP were tested to analyze their effect in

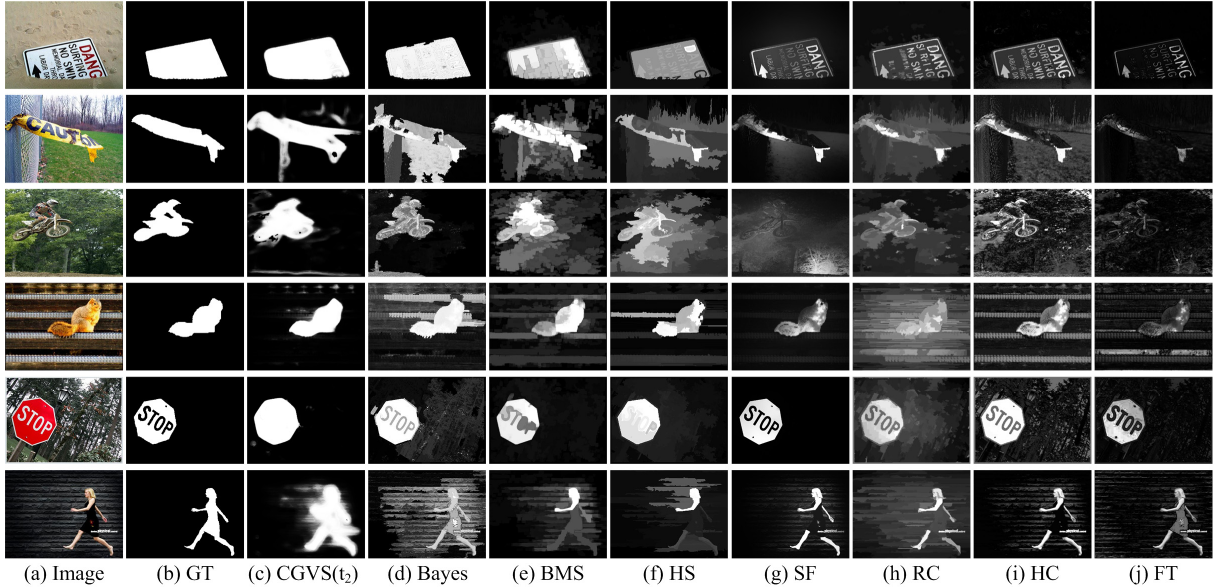

(a) Image   (b) GT   (c) CGVS($t_2$)   (d) Bayes   (e) BMS   (f) HS   (g) SF   (h) RC   (i) HC   (j) FT

Fig. 13. Visual comparison of salient object detection. (a) original images, (b) human-marked ground truth, results of salient object detection with various methods: (c) the proposed method with two steps of iteration, (d) Bayesian saliency (Bayes) [5], (e) Boolean map (BMS)[7], (f) Hierarchical saliency (HS) [6], (g) Saliency filters (SF)[14], (h) histogram-based contrast (HC) [13], (i) region-based contrast (RC) [13], and (j) Frequency-tuned (FT)[2].

this experiment. We tested the robustness of our method to the parameters of $d_r$ and $\sigma_c$ in the range of $\{1/2, 1/3, 1/4, 1/5, 1/6\} \cdot min(W, H)$ used in Section 3.1. Fig. 14 shows the F-scores and weighted F-score on the whole salient dataset (ASD) when varying these parameters. It can be seen that our system is very robust to these parameters. The same conclusion can be drawn with fixation prediction experiments on Judd dataset (not shown here for space limitation).

**5.5. Salient Structure Detection Using Saliency Map**
Fixation prediction methods usually obtain poor performance when facing the task of salient object detection. This is understandable because almost all fixation prediction methods ignore the object surface and shape information. In this study, we try to bridge the gap between the tasks of fixation prediction and salient object detection by transforming saliency map to salient structure. Fig. 15 shows that the performances of several fixation prediction methods are significantly improved for salient object detection on both metrics of P-R Curve and weighted F-score.

**6. Discussion and Conclusion**
In this paper, we proposed a context-guided visual search (CGVS) system based on the guided search theory. Different from the classical FIT theory [19] and the popular model proposed by Itti *et al*. [15], our method searches salient structures with Bayesian inference guided by context information, such as the location and size of salient structure, importance of feature, etc. This property endows our system the capability of adapting to various scenes. As a result, our method achieves a competitive performance on the both tasks of fixation prediction and salient object detection comparing to the state-of-the-art methods.

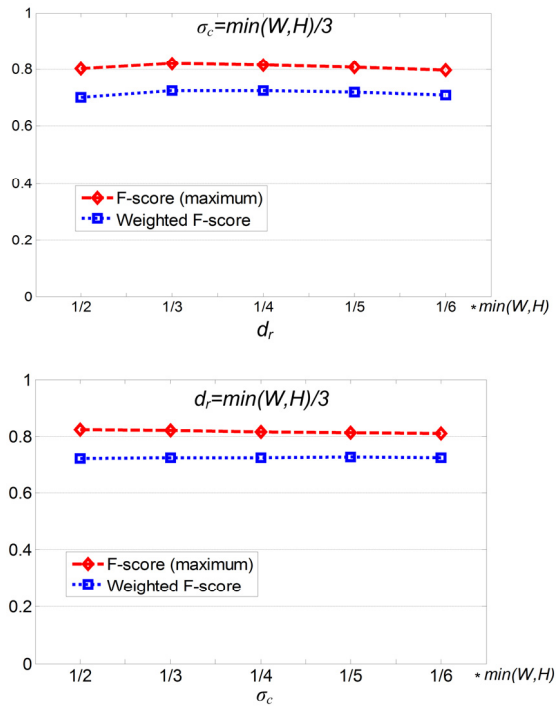

Fig. 14. Robustness to parameters on salient object dataset (ASD). **Top**: Testing our CGVS($t_2$) with various $d_r$ and $\sigma_c = min(W, H) / 3$. **Bottom**: Testing our CGVS($t_2$) with various $\sigma_c$ and $d_r = min(W, H) / 3$.

It is worth to note that the proposed model can be regarded as a unified and general framework for guided search. For example, this system is easy to be extended by introducing high-level object-related global features in the non-selective pathway for a specific object search task. Meanwhile, adding more local features (e.g., depth, motion, etc.) in the selective pathway could also extend our system for more applications such as video processing.

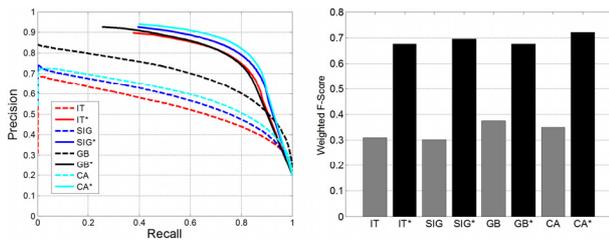

Fig. 15. Evaluating the performances of several fixation prediction methods and their modified versions (*) for salient structure detection with both metrics of P-R Curve (**Left**) and weighted F-score (**Right**).

**Acknowledgements**

This work was supported by the 973 Project under Grant 2013CB329401, the NSFC under Grant 61375115, 91420105, and the Doctoral Support Program of UESTC.

**References**

[1] T. Judd, K. Ehinger, F. Durand, and A. Torralba, "Learning to predict where humans look," in *IEEE International Conference on Computer Vision*, 2009, pp. 2106-2113.

[2] R. Achanta, S. Hemami, F. Estrada, and S. Susstrunk, "Frequency-tuned salient region detection," in *Computer Vision and Pattern Recognition, IEEE Conference on*, 2009, pp. 1597-1604.

[3] J. M. Wolfe, M. L. H. Võ, K. K. Evans, and M. R. Greene, "Visual search in scenes involves selective and nonselective pathways," *Trends in cognitive sciences,* vol. 15, pp. 77-84, 2011.

[4] R. Achanta, F. Estrada, P. Wils, and S. Süsstrunk, "Salient region detection and segmentation," in *International Conference on Computer Vision Systems, LNCS*, 2008, pp. 66-75.

[5] Y. Xie, H. Lu, and M.-H. Yang, "Bayesian saliency via low and mid level cues," *Image Processing, IEEE Transactions on,* vol. 22, pp. 1689-1698, 2013.

[6] Q. Yan, L. Xu, J. Shi, and J. Jia, "Hierarchical saliency detection," in *Computer Vision and Pattern Recognition (CVPR), IEEE Conference on*, 2013, pp. 1155-1162.

[7] J. Zhang and S. Sclaroff, "Saliency detection: A boolean map approach," in *Computer Vision (ICCV), 2013 IEEE International Conference on*, 2013, pp. 153-160.

[8] J. Li, M. D. Levine, X. An, X. Xu, and H. He, "Visual saliency based on scale-space analysis in the frequency domain," *Pattern Analysis and Machine Intelligence, IEEE Transactions on,* vol. 35, pp. 996-1010, 2013.

[9] Y. Zhai and M. Shah, "Visual attention detection in video sequences using spatiotemporal cues," in *Proceedings of the 14th annual ACM international conference on Multimedia*, 2006, pp. 815-824.

[10] S. Goferman, L. Zelnik-Manor, and A. Tal, "Context-aware saliency detection," *Pattern Analysis and Machine Intelligence, IEEE Transactions on,* vol. 34, pp. 1915-1926, 2012.

[11] J. Harel, C. Koch, and P. Perona, "Graph-based visual saliency," in *Advances in neural information processing systems*, 2006, pp. 545-552.

[12] C. Christopoulos, A. Skodras, and T. Ebrahimi, "The JPEG2000 still image coding system: an overview," *Consumer Electronics, IEEE Transactions on,* vol. 46, pp. 1103-1127, 2000.

[13] M.-M. Cheng, G.-X. Zhang, N. J. Mitra, X. Huang, and S.-M. Hu, "Global contrast based salient region detection," in *Computer Vision and Pattern Recognition (CVPR), IEEE Conference on*, 2011, pp. 409-416.

[14] F. Perazzi, P. Krahenbuhl, Y. Pritch, and A. Hornung, "Saliency filters: Contrast based filtering for salient region detection," in *Computer Vision and Pattern Recognition (CVPR), IEEE Conference on*, 2012, pp. 733-740.

[15] L. Itti, C. Koch, and E. Niebur, "A model of saliency-based visual attention for rapid scene analysis," *IEEE Transactions on pattern analysis and machine intelligence,* vol. 20, pp. 1254-1259, 1998.

[16] U. Rutishauser, D. Walther, C. Koch, and P. Perona, "Is bottom-up attention useful for object recognition?," in *Computer Vision and Pattern Recognition, IEEE Conference on*, 2004, pp. II-37-II-44 Vol. 2.

[17] N. Bruce and J. Tsotsos, "Saliency based on information maximization," in *Advances in neural information processing systems*, 2005, pp. 155-162.

[18] X. Hou, J. Harel, and C. Koch, "Image signature: Highlighting sparse salient regions," *Pattern Analysis and Machine Intelligence, IEEE Transactions on,* vol. 34, pp. 194-201, 2012.

[19] A. M. Treisman and G. Gelade, "A feature-integration theory of attention," *Cognitive psychology,* vol. 12, pp. 97-136, 1980.

[20] C. Koch and S. Ullman, "Shifts in selective visual attention: towards the underlying neural circuitry," in *Matters of intelligence*, ed: Springer, 1987, pp. 115-141.

[21] L. Zhang, M. H. Tong, T. K. Marks, H. Shan, and G. W. Cottrell, "SUN: A Bayesian framework for saliency using natural statistics," *Journal of vision,* vol. 8, p. 32, 2008.

[22] A. Garcia-Diaz, X. R. Fdez-Vidal, X. M. Pardo, and R. Dosil, "Decorrelation and distinctiveness provide with human-like saliency," in *Advanced concepts for intelligent vision systems*, 2009, pp. 343-354.

[23] L. Itti and C. Koch, "Computational modeling of visual attention," *Nature reviews neuroscience,* vol. 2, pp. 194-203, 2001.

[24] J. M. Wolfe, "Guided search 2.0 a revised model of visual search," *Psychonomic bulletin & review,* vol. 1, pp. 202-238, 1994.

[25] A. Borji and L. Itti, "State-of-the-art in visual attention modeling," *Pattern Analysis and Machine Intelligence, IEEE Transactions on,* vol. 35, pp. 185-207, 2013.

[26] A. Borji, M.-M. Cheng, H. Jiang, and J. Li, "Salient object detection: A survey," *arXiv preprint arXiv:1411.5878,* 2014.

[27] O. Le Meur, P. Le Callet, D. Barba, and D. Thoreau, "A coherent computational approach to model bottom-up visual attention," *Pattern Analysis and Machine Intelligence, IEEE Transactions on,* vol. 28, pp. 802-817, 2006.

[28] A. Borji and L. Itti, "Exploiting local and global patch rarities for saliency detection," in *Computer Vision and Pattern Recognition (CVPR), 2012 IEEE Conference on*, 2012, pp. 478-485.

[29] X. Hou and L. Zhang, "Saliency detection: A spectral residual approach," in *Computer Vision and Pattern Recognition, 2007. CVPR'07. IEEE Conference on*, 2007, pp. 1-8.
[30] C. Guo and L. Zhang, "A novel multiresolution spatiotemporal saliency detection model and its applications in image and video compression," *Image Processing, IEEE Transactions on,* vol. 19, pp. 185-198, 2010.
[31] C. Guo, Q. Ma, and L. Zhang, "Spatio-temporal saliency detection using phase spectrum of quaternion fourier transform," in *Computer vision and pattern recognition, 2008. cvpr 2008. ieee conference on*, 2008, pp. 1-8.
[32] A. Borji, "Boosting bottom-up and top-down visual features for saliency estimation," in *Computer Vision and Pattern Recognition (CVPR), 2012 IEEE Conference on*, 2012, pp. 438-445.
[33] W. Einhäuser, M. Spain, and P. Perona, "Objects predict fixations better than early saliency," *Journal of vision,* vol. 8, p. 18, 2008.
[34] L. Elazary and L. Itti, "Interesting objects are visually salient," *Journal of vision,* vol. 8, p. 3, 2008.
[35] M. Cerf, J. Harel, W. Einhäuser, and C. Koch, "Predicting human gaze using low-level saliency combined with face detection," in *Advances in neural information processing systems*, 2008, pp. 241-248.
[36] S. Ramanathan, H. Katti, N. Sebe, M. Kankanhalli, and T.-S. Chua, "An eye fixation database for saliency detection in images," in *Computer Vision–ECCV 2010*, ed: Springer, 2010, pp. 30-43.
[37] A. Oliva, A. Torralba, M. S. Castelhano, and J. M. Henderson, "Top-Down Control of Visual Attention in Object Detection," in *Image Processing, 2003. ICIP 2003. Proceedings. 2003 International Conference on*, 2003, pp. I - 253-6.
[38] V. Navalpakkam and L. Itti, "Modeling the influence of task on attention," *Vision Research,* vol. 45, pp. 205–231, 2005.
[39] V. Navalpakkam and L. Itti, "An Integrated Model of Top-Down and Bottom-Up Attention for Optimizing Detection Speed," in *2013 IEEE Conference on Computer Vision and Pattern Recognition*, 2006, pp. 2049-2056.
[40] R. J. Peters and L. Itti, "Beyond bottom-up: Incorporating task-dependent influences into a computational model of spatial attention," in *Computer Vision and Pattern Recognition, 2007. CVPR'07. IEEE Conference on*, 2007, pp. 1-8.
[41] Q. Zhao and C. Koch, "Learning a saliency map using fixated locations in natural scenes," *Journal of vision,* vol. 11, p. 9, 2011.
[42] W. Kienzle, F. Wichmann, B. Schölkopf, and M. Franz, "A nonparametric approach to bottom-up visual saliency," *Advances in Neural Information Processing Systems,* 2006S.
[43] A. Torralba, A. Oliva, M. S. Castelhano, and J. M. Henderson, "Contextual guidance of eye movements and attention in real-world scenes: The role of global features in object search," *Psychological Review,* vol. 113, pp. 766--786, 2006.
[44] A. Torralba, "Modeling global scene factors in attention," *J Opt Soc Am A Opt Image Sci Vis,* vol. 20, pp. 1407--1418, 2003.
[45] L. Itti and P. F. Baldi, "Bayesian surprise attracts human attention," in *Advances in neural information processing systems*, 2005, pp. 547-554.
[46] T. Liu, J. Sun, N.-n. Zheng, X. Tang, and H.-y. Shum, "Learning to detect a salient object," in *Computer and Vision Pattern Recognition, IEEE Conference on* 2007, pp. 1-8.
[47] R. Margolin, A. Tal, and L. Zelnik-Manor, "What makes a patch distinct?," in *Computer Vision and Pattern Recognition (CVPR), 2013 IEEE Conference on*, 2013, pp. 1139-1146.
[48] D. A. Klein and S. Frintrop, "Center-surround divergence of feature statistics for salient object detection," in *Computer Vision (ICCV), 2011 IEEE International Conference on*, 2011, pp. 2214-2219.
[49] W. Zhu, S. Liang, Y. Wei, and J. Sun, "Saliency optimization from robust background detection," in *Computer Vision and Pattern Recognition (CVPR), 2014 IEEE Conference on*, 2014, pp. 2814-2821.
[50] Y. Wei, F. Wen, W. Zhu, and J. Sun, "Geodesic saliency using background priors," in *Computer Vision–ECCV 2012*, ed: Springer, 2012, pp. 29-42.
[51] V. Ferrari, T. Deselaers, and B. Alexe, "Measuring the Objectness of Image Windows," *Pattern Analysis and Machine Intelligence, IEEE Transactions on,* vol. 34, pp. 2189-2202, 2012.
[52] I. Endres and D. Hoiem, "Category-Independent Object Proposals with Diverse Ranking," *Pattern Analysis and Machine Intelligence, IEEE Transactions on,* vol. 36, pp. 222 - 234, 2014.
[53] M.-M. Cheng, Z. Zhang, W.-Y. Lin, and P. Torr, "BING: Binarized Normed Gradients for Objectness Estimation at 300fps," in *Computer Vision and Pattern Recognition (CVPR), 2014 IEEE Conference on*, 2014, pp. 3286 - 3293.
[54] Y. Li, X. Hou, C. Koch, J. Rehg, and A. Yuille, "The secrets of salient object segmentation," in *Computer Vision and Pattern Recognition (CVPR), IEEE Conference on*, 2014.
[55] M. Holtzman-Gazit, L. Zelnik-Manor, and I. Yavneh, "Salient edges: A multi scale approach," in *ECCV, Workshop on Vision for Cognitive Tasks*, 2010.
[56] K. Yang, S. Gao, C. Li, and Y. Li, "Efficient color boundary detection with color-opponent mechanisms," in *Computer Vision and Pattern Recognition (CVPR), 2013 IEEE Conference on*, 2013, pp. 2810-2817.
[57] B. W. Tatler, "The central fixation bias in scene viewing: Selecting an optimal viewing position independently of motor biases and image feature distributions," *Journal of vision,* vol. 7, p. 4, 2007.
[58] A. Borji, H. R. Tavakoli, D. N. Sihite, and L. Itti, "Analysis of scores, datasets, and models in visual saliency prediction," in *Computer Vision (ICCV), 2013 IEEE International Conference on*, 2013, pp. 921-928.
[59] R. Margolin, L. Zelnik-Manor, and A. Tal, "How to Evaluate Foreground Maps," in *Computer Vision and Pattern Recognition (CVPR), IEEE Conference on*, 2014, pp. 248-255.
[60] M.-M. Cheng, J. Warrell, W.-Y. Lin, S. Zheng, V. Vineet, and N. Crook, "Efficient salient region detection with soft image abstraction," in *Computer Vision (ICCV), 2013 IEEE International Conference on*, 2013, pp. 1529-1536.